%% file: dictmlm.tex
\documentclass[11pt,a4paper]{article}
\usepackage[hyperref]{emnlp2020}
\usepackage{times}
\usepackage{latexsym}
\usepackage{graphicx}
\usepackage{tabularx}
\usepackage{mathtools}
\usepackage{booktabs}
\usepackage{url}
\usepackage{longtable}
\usepackage{tabu}
\usepackage{multirow}
\usepackage{subfigure}

\usepackage[russian,greek,english]{babel}

\usepackage{microtype}

\aclfinalcopy 

\title{\textsc{dict-mlm}: Improved Multilingual Pre-Training using Bilingual Dictionaries.}

\author{Aditi Chaudhary\textsuperscript{1 \Thanks{ Work done as part of Google Research Internship.}}, Karthik Raman\textsuperscript{2}, Krishna Srinivasan\textsuperscript{2}, Jiecao Chen\textsuperscript{2}\\
  \textsuperscript{1}Language Technologies Institute, Carnegie Mellon University \\
  \textsuperscript{2} Google Research, Mountain View \\
\small{ \texttt{aschaudh@cs.cmu.edu}} \hspace{.5cm}
  { \texttt{\{karthikraman,krishnaps,chenjiecao\}@google.com}}}

\date{}

\begin{document}
\maketitle

\input{abstract.tex}

\input{introduction.tex}

\input{methodology.tex}

\input{experiments.tex}

\input{relatedword.tex}

\input{conclusion.tex}

 \section*{Acknowledgements}
 We would like to thank Melvin Johnson for the interesting discussions which helped shaped this work.
\bibliographystyle{acl_natbib}
\bibliography{dictmlm}

\input{appendix}

\end{document}

%% file: abstract.tex
\begin{abstract}
Pre-trained multilingual language models such as mBERT \cite{devlin-etal-2019-bert} have shown immense gains for several natural language processing (NLP) tasks, especially in the \emph{zero-shot} cross-lingual setting.
Most, if not all, of these pre-trained models rely on the masked-language modeling (MLM) objective as the key language learning objective.
The principle behind  these approaches is that predicting the masked words with the help of the surrounding text helps learn potent contextualized representations.
Despite the strong representation learning capability enabled by MLM, we demonstrate an inherent limitation of MLM for multilingual representation learning.
In particular, by requiring the model to predict the \emph{language-specific} token, the MLM objective disincentivizes learning a language-agnostic representation -- which is a key goal of multilingual pre-training.
Therefore to encourage better cross-lingual representation learning we propose the \textsc{dict-mlm} method.
\textsc{dict-mlm} works by incentivizing the model to be able to predict not just the original masked word, but potentially any of its cross-lingual synonyms as well.
Our empirical analysis on multiple downstream tasks spanning 30+ languages, demonstrates the efficacy of the proposed approach and its ability to learn better multilingual representations.
\end{abstract}

%% file: introduction.tex
\section{Introduction}
Masked Language Modeling (MLM), introduced by \citet{devlin-etal-2019-bert}, \citet{ taylor1953cloze}, in conjunction with transformers \cite{vaswani2017attention}, has spurred a rapid succession of massive pre-trained models -- RoBERTa \cite{liu1907roberta}, ALBERT \cite{lan2019albert}, ELECTRA \cite{clark2020electra}, etc.
These new models have shown significant advances in several natural language processing (NLP) tasks and set new benchmarks.
The success of MLM is attributed to the model learning better contextualized representations by virtue of using the bi-directional context to predict randomly masked tokens or sub-words in a sequence.  

MLM and its variants have also been applied successfully to the multilingual setting (mBERT\footnote{https://github.com/google-research/bert/blob/master/multilingual.md}, XLM; \citet{lample2019cross}) wherein a single model is pre-trained on the concatenated monolingual corpora of multiple languages. This has greatly improved the  cross-lingual generalizability of the pre-trained model resulting in significant improvements, especially in the \textit{zero-shot} setting where the pre-trained model is fine-tuned for the task in question on one language and directly applied to another language. 

The key to successful cross-lingual transfer learning (especially in the zero-shot setting) is the ability to learn semantically rich language-agnostic representations \cite{ruder:2019}.
The more language-agnostic a representation, the better its chances of carrying over knowledge seamlessly across languages and overcome the lack of training data in other languages \cite{upadhyay-etal-2016-cross, ruder:2019}.

However, the design of MLM is not conducive for learning language-agnostic representations.
This is because MLM requires the model to predict the masked token in the specific language, and not any of its synonyms  -- either within or across languages.
For example, in the following sentence \emph{Food and water are neccessities of life}, with the word \emph{life} masked, the MLM objective drives the model to learn 
this exact word (\emph{life}) in the given context.
This thus forces the learned representation of the word to be different from its synonyms in other languages (e.g. \emph{vida, Leben, jeevan, \ldots}).
In other words the learned representations of the model are forced to be language-specific to help identify the exact language of the sentence.
This is partially the reason behind mBERT being very good at the Language-Identification task\footnote{In fact, prior work \cite{conneau2017word} has explicitly tried to learn multilingual representations by adversarially training the model to not be able to distinguish language.} \cite{pires2019multilingual}.

This behavior is also seen when we evaluate the language-agnosticity of the learnt mBERT representations across the multiple layers of the model.
As seen in Figure~\ref{fig:tatoeba_cosine}, the 8th layer of mBERT is 25+\% better at cross-lingual semantic retrieval compared to the final layer of BERT.
This behavior has also been observed in prior work \cite{hu2020xtreme, pires2019multilingual}.
One approach proposed to improve this language-agnosticity
is translation language modeling -- TLM -- \cite{lample2019cross}.
However this relies on the use of expensive, sentence-aligned parallel corpora, while still retaining the same MLM-incentive issues.



To address these limitations, we propose \textsc{dict-mlm} which facilitates cross-lingual alignment more directly.
\textsc{dict-mlm} achieves this by incentivizing the model to be able to predict not only the masked token in its original language, but also its synonyms in other languages.
While some existing approaches rely on millions of expensive translations, we show that even with only thousands of cross-lingual word pairs (from bilingual dictionaries), \textsc{dict-mlm} is able to learn a more potent language-agnostic representation than mBERT.

We demonstrate this empirically on multiple downstream tasks including sequence labeling (NER, POS), classification (MLDOC-Headline, PAWS-X), textual entailment (XNLI) and cross-lingual sentence retrieval (TATOEBA) -- leveraging a data setup similar to the XTREME \cite{hu2020xtreme} benchmark which covers 40 typologically diverse languages.  Our contributions can be summarized as follows:
\begin{itemize}
\itemsep0em 
    \item We propose a new pre-training method \textsc{dict-mlm} to enforce cross-lingual alignment more directly. We find that \textsc{dict-mlm} is especially effective for sequence-labeling tasks (NER, POS) since these tasks rely more on token-level representations rather than the full sequence representation which is more conducive to our cross-lingual word alignment objective (Table \ref{tab:main}). 
    
    \item Overall, we observe that our model outperforms mBERT for all tasks with improvements of +2.4 F1 for NER, +3.7 accuracy for POS, +0.5 accuracy for XNLI, +2.5 accuracy for MLDOC-Headline and +5.3 accuracy for PAWS-X, averaged across all languages. Furthermore, our model also outperforms existing work that leverage parallel text which is a more expensive resource than bilingual dictionaries (Table \ref{tab:main} and Table \ref{tab:main_subset}). 
    \item Our learned representations are empirically shown to be far more cross-lingual, with 20+\% improvements observed on TATOEBA. (Fig~\ref{fig:tatoeba_cosine}).
    \item  Qualitative analysis reveals that even languages which do not have publicly available bilingual dictionaries benefit from \textsc{dict-mlm} (Table \ref{tab:dict}). 
\end{itemize}

%% file: methodology.tex
\begin{figure*}
    \centering
    \includegraphics[width=\textwidth]{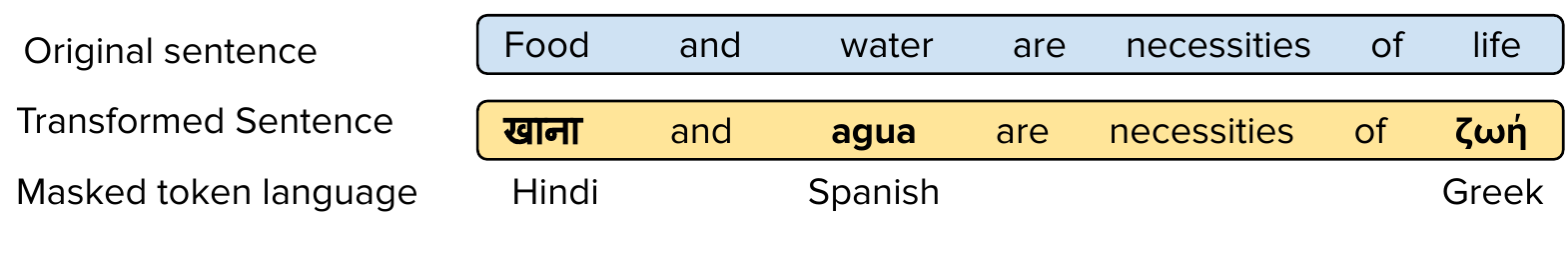}
    \caption{Generation of training data for \textsc{DICT-MLM}. Tokens in bold are masked out and \textsc{DICT-MLM} is trained to predict the corresponding cross-lingual synonym as shown in the transformed sentence. }
    \label{fig:example}
\end{figure*} 

\section{Methodology}

\textbf{Our Focus}: While there exists different (often complementary) techniques to improve cross-lingual transfer learning -- including use of larger, richer pretraining datasets \cite{conneau2020unsupervised} or finetuning using data augmentation \cite{fang2020filter} -- our focus is specifically on the improving the pretraining objective that is common to all of these approaches, namely MLM.

In order to address the aforementioned limitations of MLM, we propose \textsc{DICT-MLM}  which leverages bilingual dictionaries  during pre-training to explicitly facilitate  cross-lingual word alignment.  
In Section \S \ref{sec:algorithm} we describe the training data creation for \textsc{DICT-MLM} followed by the model and training regimen in Section \S \ref{sec:training}.
\subsection{Preliminaries}
\label{sec:mlm}
Given an input text stream $W\!=\!w_1, w_2 \cdots w_n$,  MLM uses the past and future tokens $W_{\setminus t}\!:=\!(w_1,\cdots, w_{t-1}, w_{t+1}, \cdots, w_{n}$) to predict the masked token $w_t$. \citet{devlin-etal-2019-bert} apply this objective on the concatenated monolingual corpora of 104 languages to get a pre-trained multilingual model (mBERT).\footnote{https://github.com/google-research/bert/blob/master/multilingual.md}

\subsection{Preparing the Training Data }
\label{sec:algorithm}
To facilitate cross-lingual word alignment during pre-training,
\textsc{DICT-MLM} is trained to predict the cross-lingual synonym of the masked token.
To train this model, we first create  multilingual code-switched sentences from the monolingual corpora of $L$ languages by leveraging bilingual dictionaries.
Consider the monolingual corpus $\mathcal{D}_l=\{\mathbf{x_1^l}, \mathbf{x_2^l}, \cdots, \mathbf{x_n^l} \}$ for language \textit{l}. For each sentence $\mathbf{x_i^l} \in \mathcal{D}_l$, we randomly select (and mask) 15\% of whole words which have at least one dictionary entry in any bilingual lexicon.
For each such masked token ($w^l \in \mathbf{x_i^l}$), we first retrieve all its synonyms from the bilingual dictionaries\footnote{Some words occur in multiple bilingual dictionaries thus allowing us to potentially extract synonyms across multiple languages in these cases.}. From the retrieved set of candidate (cross-lingual) synonyms  we then randomly select one synonym given by $s_w^{l_d}$ where $l_d$ refers to the language of the selected synonym.
An example of such a generated sentence can be seen in Figure \ref{fig:example}.
Note that since synonyms for masked tokens are sampled independently, the resulting sentences could contain tokens from multiple different languages (as seen in above example).
This makes the masking task a lot more challenging, as
the model not only needs to learn the semantics of the word in the masked slot, but also needs to predict the word in the desired language (which may not be the same as the surrounding context).
Our hypothesis is that this dual-challenge now forces the learned representations to be both semantically and cross-lingually rich.
By repeating this process for all monolingual corpora of the $L$ different languages, we get the desired training dataset\footnote{Note that in our experiments, we used bilingual lexicons for only 45 languages. Languages without any lexicon instead relied only on vanilla MLM instead.}. 

After generating this multilingual code-switched data we perform WordPiece \cite{wu2016google} tokenization following BERT \cite{devlin-etal-2019-bert}. In the original implementation of BERT \cite{devlin-etal-2019-bert}, the authors mask  individual word-pieces within a sentence which however causes only some word-pieces within a word to be masked making MLM a simpler task. The authors later proposed to perform whole-word masking (WWM) wherein all the wordpieces of a word are masked while keeping the percentage of masked tokens as before. They empirically  find WWM to perform better hence we  also perform WWM for \textsc{DICT-MLM}. We further use a \emph{dynamic masking} strategy, as proposed by RoBERTa \cite{liu1907roberta}, which duplicates the data $n$ times causing different tokens to be selected for masking each time. For \textsc{DICT-MLM}, this amplifies the randomness since during the dictionary-replacement step the cross-lingual synonym is also selected randomly.

 \begin{figure*}[h]
     \centering
     \includegraphics[width=\textwidth]{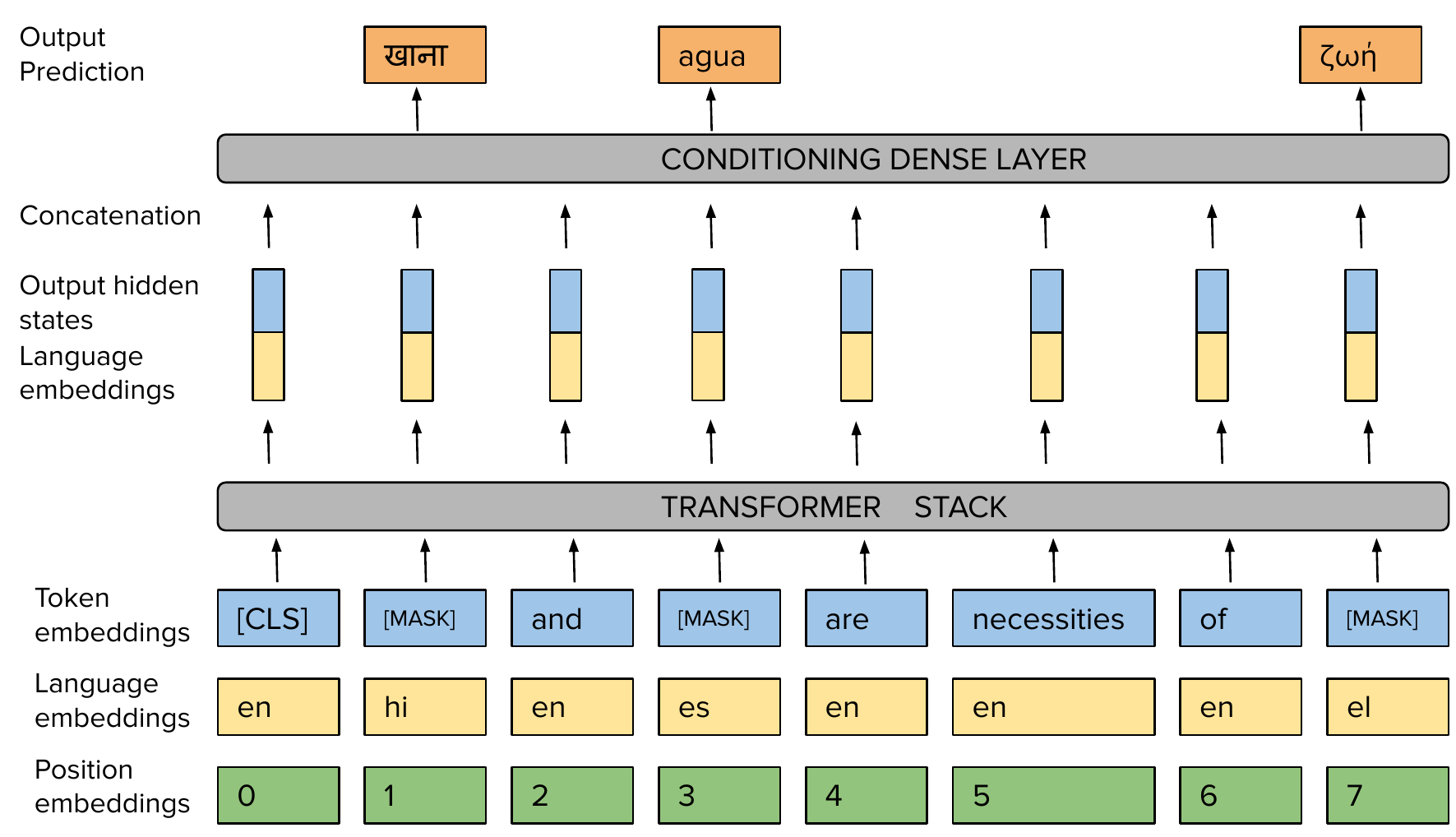}
     \caption{Illustration of our \textsc{DICT-MLM} model for the input sentence `Food and water are necessities of life'.}
     \label{fig:model}
 \end{figure*}
\subsection{Model and Training Regimen }
\label{sec:training}
We next describe the training setup used for pre-training the different models. We base our model on mBERT \cite{devlin-etal-2019-bert} which uses a transformer stack \cite{vaswani2017attention} and is trained on the Wikipedia monolingual corpora of 104 languages. For the \textsc{DICT-MLM} methods, we did not use the next sentence prediction (NSP) objective as prior work \cite{liu1907roberta} has shown that to add very little gains.

As in the original mBERT implementation, masked tokens are replaced with the [MASK] token 80\% of the time, 10\%  of the time with the original token and the remaining 10\% with a random token.
Given that downstream tasks/datasets are typically monolingual in nature, we combine our \textsc{DICT-MLM} objective with the regular MLM objective to ensure the model learns from monolingual contexts just as well.
In particular, we multi-task these two objectives by setting the label of the masked token to be the cross-lingual synonym $t$\% of the time.
For the remaining (100-$t$\%) time the original token is used as the label.
We show the effect of the value of $t$ in our experimental section.


At this point, an astute reader may wonder how a model can learn to predict masked tokens in different languages without being given any cue about the target token's language.
To remedy this we make two architectural changes to the mBERT model.
First, we add a language embedding layer to the input layer similar to the position and segment embeddings layer.
Furthermore, since we want to provide the MLM prediction task with additional language context to help it predict tokens in the appropriate language, we make a second change as well.
Specifically we modify the dense transformation that happens as part of the MLM prediction.
Rather than providing this dense block only the token embedding of the final transformer layer as in mBERT, we additionally concatenate this token representation with an embedding of  the associated language.
As in MLM, the output of this dense transformation is used to make a prediction over the shared wordpiece vocabulary via softmax. 
These modifications are illustrated in Figure \ref{fig:model}.
It is worth noting that the latter change -- of the additional conditioning before the MLM softmax -- only affects the pretraining and not the finetuning.

Furthermore since the number of languages is much smaller than the vocabulary size, these changes add very few parameters to the overall model, and do not affect inference / training speed.
While both of the above mentioned modifications require the use of language embeddings, which could be independently learned, in practice we found that coupling the two help improve performance slightly and is thus the setup we use in our experimentation.
We also looked to understand the effect of these architecture changes in our empirical analyses.
 
 \paragraph{\textsc{dict-tlm}} Inspired by the translation language modeling (TLM) objective \cite{lample2019cross}, we experiment with a similar TLM variant of \textsc{dict-mlm} which uses bilingual dictionaries instead of parallel data.
 Specifically, rather than rely on parallel corpora for sentence-aligned translations, we instead simulate translations using the multilingual code-switched sentences as described in Section \S \ref{sec:training}.
 As in TLM, this (synthetic) \emph{translation} is concatenated with the original sentence, with 15\% of the tokens of the concatenated sentence being masked (WWM) and predictions made using the MLM objective.
 We refer to this model variant as \textsc{dict-tlm}.

%% file: experiments.tex
\section{Experiments}
\label{sec:expt}
In this section, we empirically demonstrate the efficacy of \textsc{dict-mlm} on several downstream tasks including textual entailment, sequence labeling, sentence classification and sentence retrieval.
 \begin{table*}[h]
\centering
 \resizebox{\textwidth}{!}{
 \begin{tabular}{l|l|c|c|c|c|c|c}
 & \textbf{Model} & \multicolumn{2}{c|}{\textbf{Sequence Labeling}}& \multicolumn{2}{c|}{\textbf{Sentence Classification}} & \textbf{Textual Entailment}  & \textbf{Sentence Retrieval}\\
 & & \textbf{NER (F1)} & \textbf{POS (Acc.)} & \textbf{MLDOC-H (Acc.)} & \textbf{PAWS-X (Acc.)} & \textbf{XNLI (Acc.)}& \textbf{TATOEBA (Acc.)}\\
  \toprule
 \multirow{3}{*}{Baseline} &  \textsc{mBERT (ours)} & 68.9   & 67.9  & 65.2   & 79.5    & 68.1 & 33.3\\
                        & \textsc{mBERT (public) }& 62.2 &71.5 &  - & 81.9 & 65.4 & 38.7 \\
                        & \textsc{XLM} & 61.2 & 71.3 & - & 80.9 & 69.1 & 32.6 \\ 
                         
 \midrule
 \multirow{3}{*}{Ours} &  \textsc{dict-mlm-50}  & \textbf{71.3 $\pm_{0.5}$} & 70.4 $\pm_{0.1}$  & 66.1 $\pm_{1.9}$  & 84.2 $\pm_{0.4}$ & 67.9 $\pm_{2.5}$ & 46.4   \\
                       &  \textsc{dict-mlm-70} & 69.0 $\pm_{0.1}$ & 69.9 $\pm_{0.2}$ & 66.1 $\pm_{1.0}$ & 84.6   & \textbf{68.6 $\pm_{1.1}$}  & \textbf{47.3}  \\
                     &  \textsc{dict-mlm-90} & 67.8 $\pm_{0.6}$  & 69.4 $\pm_{2.3}$  & 65.6 $\pm_{1.0}$  & 84.1 $\pm_{0.2}$  & 67.2 $\pm_{1.8}$ & 45.1   \\
                     &  \textsc{dict-tlm} & 65.0  & \textbf{71.6}  & \textbf{67.7} & \textbf{84.8} & 66.6 & 36.9  \\
 \bottomrule
 \end{tabular}
 }

 \caption{Average scores for each task across the respective languages reveal that our proposed methods outperform  existing baselines. The models/tasks for which we were able to train multiple runs, we report the mean and std. deviation across two runs, wherever applicable. We refer to XTREME \citet{hu2020xtreme} for the  \textsc{mBERT (public) } and \textsc{XLM} results. For TATOEBA, we report the accuracy averaged across last four layers while \citet{hu2020xtreme} report results by concatenating the representations from those last four layers.  }
   \label{tab:main}
 \end{table*}

\textbf{Data:} 
We use the Wikipedia monolingual corpora for 104 languages. Owing to wide differences in the monolingual data availability of the different languages we employ a temperature based data sampling policy following \citet{siddhant2020evaluating}.
We use the publicly released MUSE \cite{conneau2017word} bilingual dictionaries,  which comprises of  110 dictionaries spanning 45 (of the 104) languages.
We pre-process all 110 dictionaries to create a single dictionary aggregating all synonyms across multiple languages as seen in Table \ref{tab:data}.
While 40 of the 45 languages have dictionaries available only to/from English, five languages (German, Spanish, Italian, French, Portuguese) have additional dictionaries between them.
As mentioned in Section \S \ref{sec:training}, we randomly sample one synonym for each (\textsc{dict-mlm}) masked token  ensuring that synonyms from all languages have equal probability. 

\subsection{Experimental setup}
We follow the same hyperparameter settings as mBERT \cite{devlin-etal-2019-bert} using 12 layers of the transformer stack \cite{vaswani2017attention} with 768 hidden units. We use a learning rate of 0.0016 with the LAMB optimizer \cite{lamboptimizer}  and a batch size of 8192. We train all our models including the mBERT baseline with the above hyperparameters on 128 TPU v3 chips for 500K steps.
We use language embeddings of 768 hidden size. 
We experiment with three different \textsc{dict-mlm} models by varying the percentage of regular MLM tokens denoted by $t=[50\%, 70\%, 90\%]$, as outlined in Section \S \ref{sec:training}.
To account for model variance we pre-train the \textsc{dict-mlm} model twice and report the averaged results (along with the variance).
We believe this provides a more fair assessment of the efficacy of these deep models given their inherent variance across runs.

During fine-tuning we use a learning rate of $3e-5$ with the AdamW optimizer and train for 8 epochs. 
In order to be consistent with the pre-training setup, for all \textsc{dict-mlm} fine-tuning runs we add language embeddings with the token embeddings as input to the transformer stack.
We evaluate our model in the zero-shot setting where we fine-tune the models only on the English portion of the tasks and directly apply the model on the test portion of the different languages. 

\subsection{Tasks}
\paragraph{Textual Entailment} We use cross-lingual natural language inference XNLI \cite{conneau2018xnli} dataset which covers 15 languages. The model takes in two input sentences and is required to classify into one of the three labels: \emph{entailment, contradiction, neutral}.

\paragraph{Sequence Labeling} We use named entity recognition NER \cite{Pan2017} and part-of-speech tagging POS \cite{nivre2018universal} datasets which cover 38 languages and 33 languages respectively. We use the same set of languages as the XTREME benchmark \cite{hu2020xtreme}. For NER, the model takes in an input sequence and is required to identify entities of the following three types: \emph{PER, ORG, LOC}. For POS tagging, the model is required to tag each token in the input sequence with its universal part-of-speech (UPOS) tag.

\paragraph{Sentence classification} We use the PAWS-X \cite{Yang2019paws-x} dataset which takes two input sequences and classifies whether these sequences are paraphrases of each other. This is available for 7 languages. We also use the MLDOC \cite{mldoc} dataset which performs document classification (into 4 categories) given an input headline and is available for 8 languages. 
 
 \paragraph{Sentence Retrieval} We use the TATOEBA \cite{Artetxe2019massively} dataset which contains upto 1000 English-aligned sentences across 122 languages. We use the same set of languages as the XTREME benchmark \cite{hu2020xtreme}. We find the nearest neighbor using cosine similarity of the sentence embeddings.
 Sentence embeddings are derived by mean-pooling of the token embeddings.\footnote{The results and trends are similar for other distance functions and pooling strategies but we leave out these variants for the sake of brevity.}

  \begin{table*}[h]
\centering
 \resizebox{\textwidth}{!}{
 \begin{tabular}{l|l|c|c|c}
 & \textbf{Model} & \multicolumn{2}{c|}{\textbf{Sequence Labeling}} & \textbf{Textual Entailment}  \\
 & & \textbf{NER (F1)} & \textbf{POS (Acc.)}  & \textbf{XNLI (Acc.)}  \\
  \toprule
 \multirow{3}{*}{Baseline} &  \textsc{mBERT (ours)} & 76.4 & 68.7  &  71.5 \\
                        &  \textsc{mBERT (theirs)*} & 67.7 $\pm_{1.3}$  & 78.3 $\pm_{0.5}$  & 70.1 $\pm_{0.8}$    \\
                        & \citet{wu20emnlp} & 67.1 $\pm_{1.1}$  & \textbf{79.0 $\pm_{0.7}$ }  & 70.5 $\pm_{0.7}$  \\

 \midrule
 \multirow{3}{*}{Ours} &  \textsc{dict-mlm-50}  
                       & \textbf{77.5 $\pm_{0.5}$ } & 71.7 $\pm_{0.9}$  & 71.6 $\pm_{2.3}$  \\
                       &  \textsc{dict-mlm-70} 
                       & 76.1 $\pm_{2.0}$  & 73.1 $\pm_{0.4}$  & \textbf{72.7 $\pm_{2.4}$ }  \\
                       &  \textsc{dict-mlm-90} 
                       & 75.7 $\pm_{1.1}$  & 71.6 $\pm_{2.5}$  & 72.2 $\pm_{0.9}$  \\
                       &  \textsc{dict-tlm} 
                       & 72.0   & 73.1   & 70.2  \\
 \bottomrule
 \end{tabular}
 }
 \caption{Average scores for each task across the 9 languages (as reported in \citet{wu20emnlp}) reveal that our proposed dictionary-based methods outperform the existing baseline which uses parallel data. \textsc{mBERT (theirs)*} refers to the results reported for mBERT by \citet{wu20emnlp} and \textsc{mBERT (ours)} refer to the scores from our re-trained model. For our trained models we report the mean and std. deviation across two runs, wherever applicable. }
   \label{tab:main_subset}
 \end{table*}

\subsection{Results}
 In Table \ref{tab:main} we report results for the six tasks, averaged across all languages and runs.\footnote{We use the same set of test languages as used by the XTREME \cite{hu2020xtreme} benchmark. Due to time constraints we only pre-train the \textsc{dict-mlm-x} models twice.} We find that our proposed models significantly outperform the vanilla  mBERT baselines.
 Furthermore, the \textsc{dict-mlm} models even outperform existing works that leverage (millions of) parallel sentence pairs, specifically the \textsc{XLM} model \cite{lample2019cross}.
 Table \ref{tab:main_subset} directly compares the performance of our approach versus the recently proposed contrastive alignment method \cite{wu20emnlp}.
 In particular, \citet{wu20emnlp} -- which proposes several alignment methods that use parallel data to improve the learned multilingual representations\footnote{We compare with the \textbf{Strong Align} variant from \citet{wu20emnlp} which improves upon the mBERT model in most cases.} -- only report results on NER, POS and XNLI for 9 languages.
 Table \ref{tab:main_subset} reports performance for those languages and tasks with significant gains observed for NER and XNLI tasks. Interestingly on POS tagging, \citet{wu20emnlp} report significantly higher results for their mBERT baseline itself than either we could reproduce or that previous works reported \cite{hu2020xtreme}.
 We still observed \textsc{dict-mlm} outperform (our implementation of) mBERT and leave resolving this disparity to future work.
 A further discussion on a possible explanation for \textsc{dict-mlm}'s reduced performance on POS-tagging can be found in Sec. \S \ref{sec:analysis}.
 
\begin{figure}[t]
    \centering
    \includegraphics[width=\columnwidth]{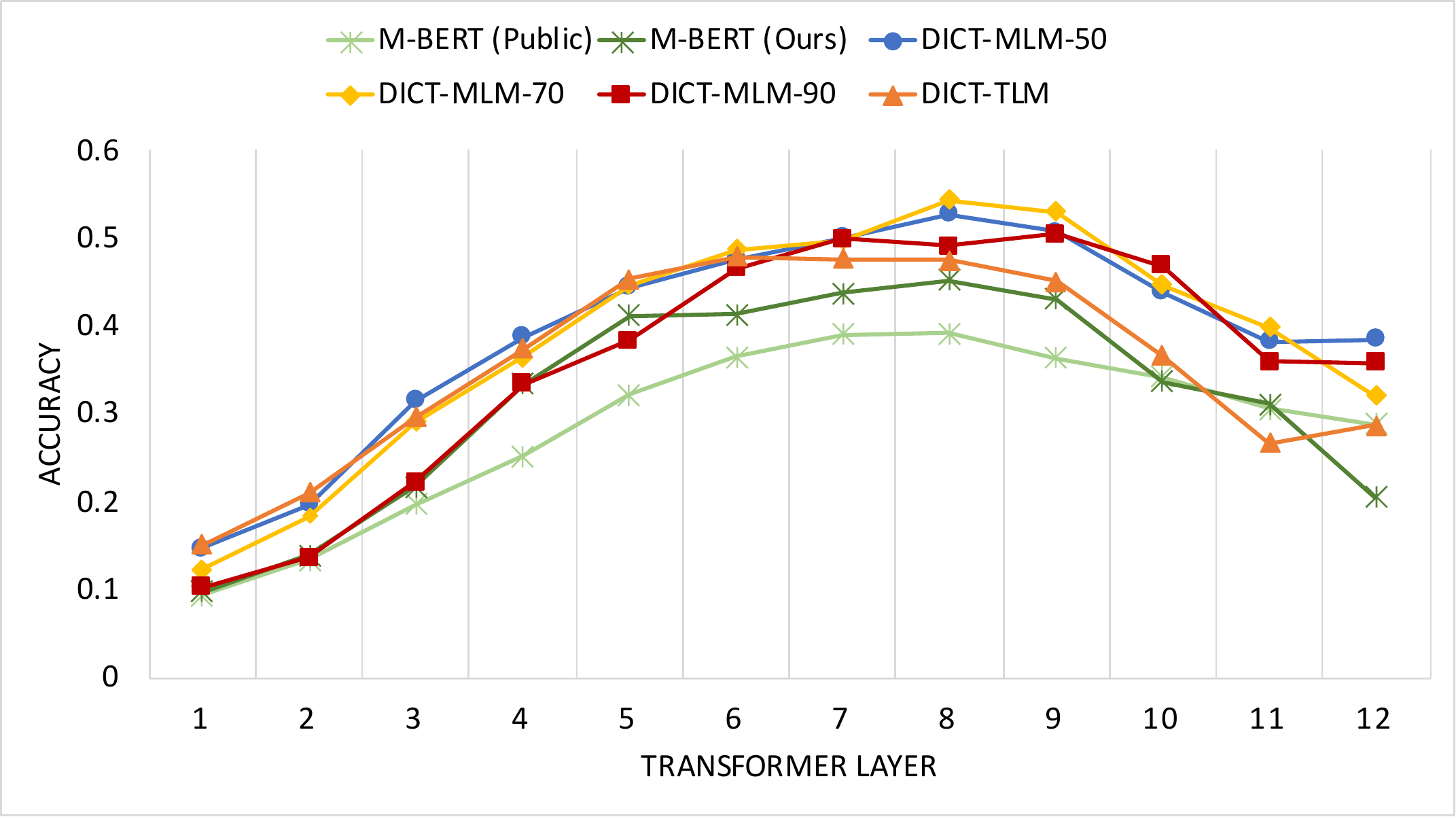}
    \caption{Accuracy of different models on the TATOEBA cross-lingual sentence retrieval task using representations from different layers of the transformer stack. The y-axis reports the nearest-neighbor accuracy and x-axis denotes the layer number. }
    \label{fig:tatoeba_cosine}
\end{figure} 

\begin{table}[ht]
 \begin{center}  \resizebox{\columnwidth}{!}{
 
 \begin{tabular}{l|l|c}
 \textbf{\textsc{task}} & \textbf{\small{\textsc{without dict}}} & \textbf{\small{\textsc{with dict}}} \\
 \toprule
NER	 & +2.3 (21) &	+2.4 (27) \\
POS &  +0.9 (6) & +3	(27)\\
TATOEBA   & +2.7 (9) & +6.95 (27) \\
 \end{tabular}
 }
 \caption{Each cell is the difference in average scores between  the \textsc{dict-mlm-50} model and the mBERT model, split based on whether the language was covered in the bilingual dictionaries or not. Even for languages not having dictionaries, we observe improvements over mBERT. For each setting, brackets denote the number of languages over which the average was computed.  
 }
 \label{tab:dict}
 \end{center}

 \end{table}
 
\begin{table*}[h]
 \begin{center}  \resizebox{!}{!}{
 
 \begin{tabular}{l|l}
 \textbf{\textsc{Source}} & \textbf{\textsc{Target}}  \\
 \toprule
pt: andar & it:camminare, es:piso, en:walking, en:walk \\
no: vokal &	en:vowels, en:vowel\\
ms: cubaan &	en:attempt, en:attempting, en:attempted, en:testing, en:attempts
 \end{tabular}
 }
 \caption{Example dictionary entries show that a given source word has multiple synonyms across different languages. The same source word also has synonyms across different morphological forms of which one form is randomly selected. The language codes are - pt: Portuguese, no: Norwegian, ms: Malay, it: Italian, es: Spanish, en: English. }
 \label{tab:data}
 \end{center}

 \end{table*}

\paragraph{Does \textsc{dict-mlm} help learn more language-agnostic representations?}
The primary motivation behind \textsc{dict-mlm} was overcoming the limitations of MLM when it comes to learning language-agnostic representations.
Thus to understand the cross-linguality of the learned representations of the \textsc{dict-mlm} models, we use the Tatoeba cross-lingual nearest-neighbor retrieval task.
This dataset directly measures and rewards the language-agnosticity of the learned representations.
As seen in Figure \ref{fig:tatoeba_cosine}, we find that our proposed methods are able to learn much better cross-lingual representations across *all* layers relative to mBERT.
Relative to the steep drop-off in performance between middle and final layers observed in mBERT models, the \textsc{dict-mlm} models are able to retain significantly more multilingual representations in the final layer representations.
One hypothesis for the observed drop in the \textsc{dict-mlm} models' final layer is the reliance on the language embedding added to the input.
We leave further investigation of this phenomena to future work.

\subsection{Analysis}
\label{sec:analysis}
In this section, we aim to understand the behavior and performance of the proposed \textsc{dict-mlm} models in more detail.

\paragraph{Is there a relation between dictionary availability and the downstream performance?} As mentioned earlier, the MUSE bilingual dictionaries cover only 45 of the 104 languages used for pre-training. Specifically, across these 45 languages, we find 34.15\% of the total Wikipedia tokens have at least one synonym available. Naturally a question arises \emph{whether only languages having dictionaries are benefited by this approach}? In Table \ref{tab:dict} we  therefore compare the downstream performance on languages which have dictionaries available versus those  which do not. We report the difference in the  (average) scores between the \textsc{dict-mlm-50} model and the mBERT model for NER, POS and TATOEBA.\footnote{For MLDOC-H, PAWS-X and XNLI tasks most or all languages are covered by the dictionaries and hence we do not report results for them.} Needless to say, languages which have dictionaries available show more improvements however, we do observe significant gains even for languages which have no dictionaries available.
In particular, the  gains on the TATOEBA dataset, indicate that the \textsc{dict-mlm} models are learning a  more cross-lingual representation for all languages, despite there being no dictionaries available for most languages. 

 \begin{figure*}[h]
\centering
\subfigure[XNLI]{
\label{xnli}%
\includegraphics[width=\columnwidth]{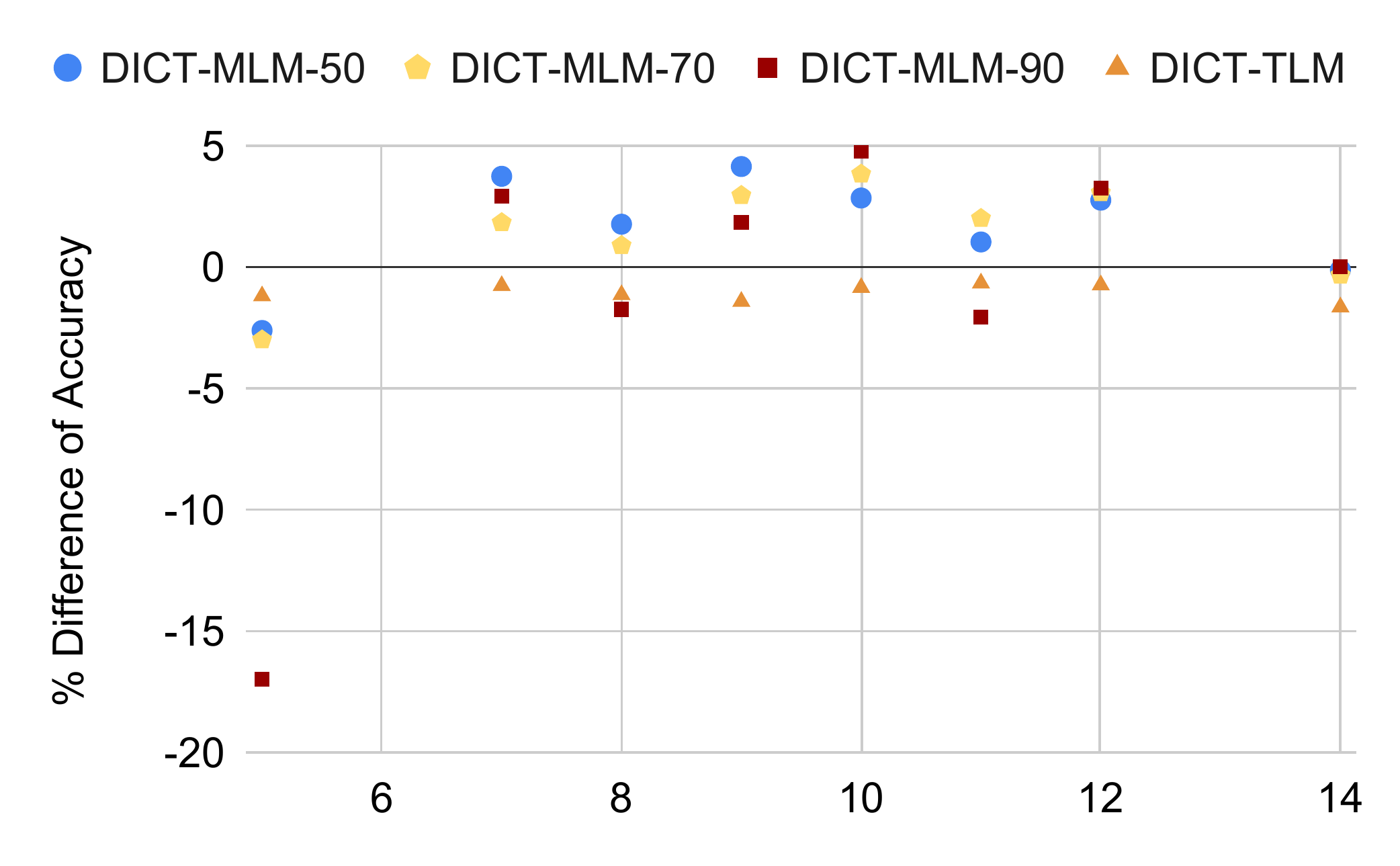}}
~
\subfigure[NER]{
\label{ner}%
\includegraphics[width=\columnwidth]{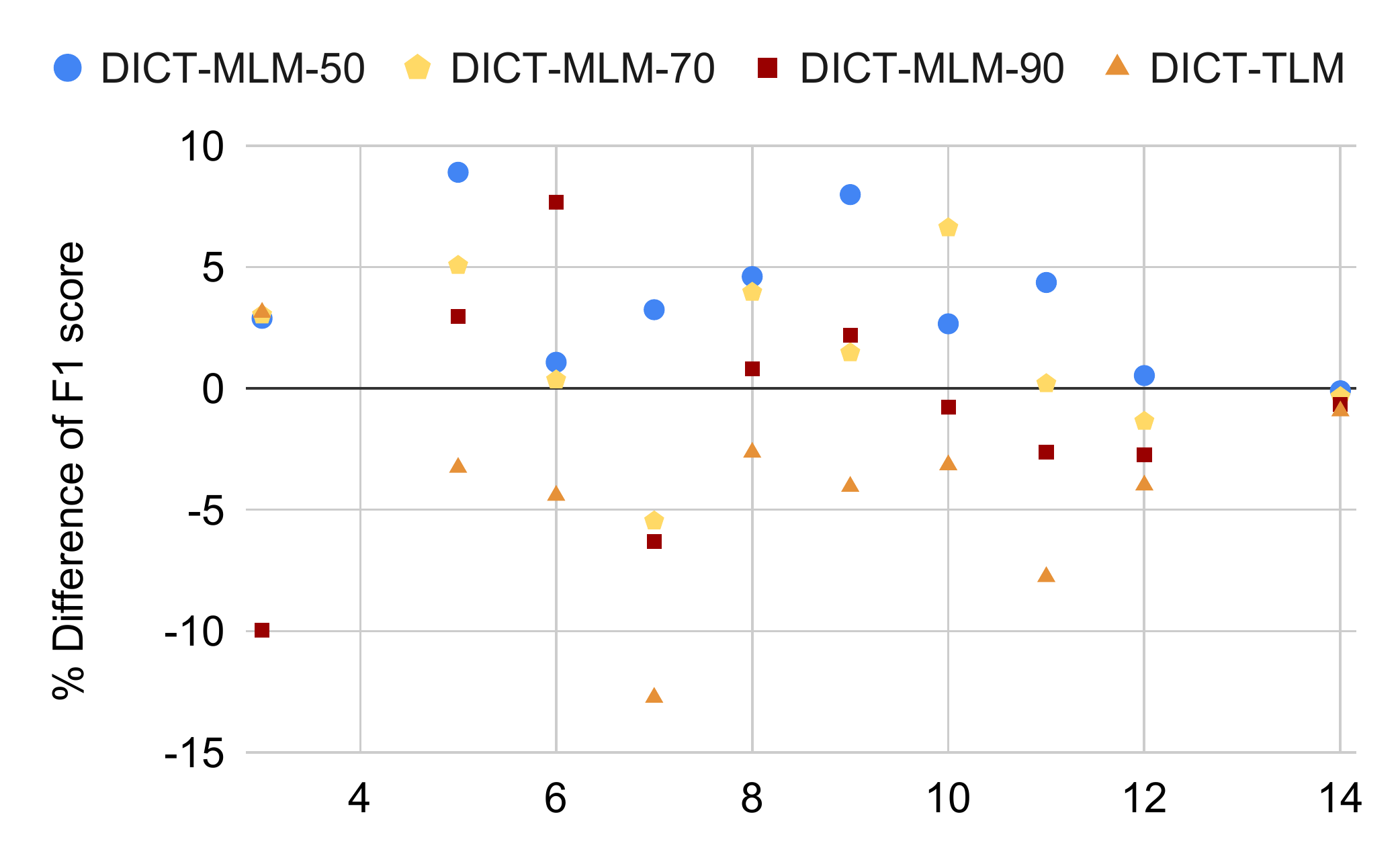}
}
~
  \caption{Comparing the (avg.) performance across the different pre-trained models grouped by Wikipedia size for a) XNLI and b) NER task. Note higher scores are better. For XNLI we plot difference in the accuracy and for NER difference in the F1 score with respect to mBERT on y-axis. }%
\label{fig:language_wiki}%
\end{figure*}

\begin{table*}[h]
 \begin{center}  \resizebox{!}{!}{
 
 \begin{tabular}{c|c|c}
 \textbf{\textsc{Task}} & \textbf{\textsc{with conditioning}} & \textbf{\textsc{without}}  \\
 \toprule
NER	& \textbf{71.3}	& 69.1 \\
POS	& 70.4	& \textbf{72.6} \\
XNLI &	\textbf{67.8} & 67.1 \\
MLDOC &	\textbf{62.1} &	\textbf{62.1} \\
PAWS-X &	84.3 &	\textbf{85.4}
 \end{tabular}
 }
 \caption{Ablation study to evaluate the effectiveness of language conditioning layer for the \textsc{dict-mlm-50} model. All scores are averaged across the respective languages. }
 \label{tab:ablation}
 \end{center}

 \end{table*}
 
\paragraph{Do some tasks/languages benefit more from  a particular \textsc{dict*} variant?}
As mentioned in Section \S \ref{sec:algorithm}, we  multi-task between the regular \textsc{mlm} objective and the \textsc{dict-mlm} objective for the three model variants.
This leads to having different amounts of cross-lingual {\it masked token} labels in the pre-training data for the variants.
In particular, 34\% of the labels of the masked tokens of the \textsc{dict-mlm-50} dataset, belong to a different language than the sentence.
Likewise this fraction is 55\% and 82\% for \textsc{dict-mlm-70} and \textsc{dict-mlm-90} respectively.\footnote{Note these numbers have been calculated on the training data which was duplicated 5 times for dynamic masking.}

Based on the results, such as those in Table~ \ref{tab:main}, we find that the \textsc{dict-mlm-50} and \textsc{dict-mlm-70} consistently score higher than the \textsc{dict-mlm-90} variant across all tasks.
This confirms our intuition for the need to multi-task with regular MLM.
By relying so heavily on predicting cross-lingual synonyms largely (82\% of masked tokens), \textsc{dict-mlm-90} does worse than the other variants on the downstream tasks which are all monolingual texts -- something it sees only 10\% of the time.
This pretraining-downstream mismatch coupled with not learning from enough monolingual examples, is likely why it trails the other two.

We also find the \textsc{dict-tlm} variant to be fairly competitive in some tasks, most notably the POS-tagging ones.
We believe this is due to a notable downside of our setup affecting the \textsc{dict-mlm} variants.
Among all of the evaluated tasks, POS tagging is the most reliant on lexical cues.
If is often the case that the internal structure (morphology) of the individual tokens is sufficient to correctly predict its POS value.
However as part of our cross-lingual synonym prediction task in the \textsc{dict-mlm} variants, we sample synonyms randomly {\emph i.e.,} we do not consider the morphology of the word or the surrounding context.
Since most publicly released dictionaries only have a single morphological form annotated, this can lead to grammatically incorrect substitutions or a loss of morphological information.
For instance, in Table \ref{tab:data} we find that the Norwegian word for  `vokal' (vowel) is mapped to both singular and plural form of the respective English word which is incorrect since Norwegian has a different word `vokaler' for the plural form (vowels). 
We posit that cleaning up the lexicons and a more careful selection of synonyms can help alleviate this issue.


We further compare the performance of the model variants with respect to the  pre-training corpus size  (\emph{wikipedia-size}) of the different languages. We group languages by their Wikipedia size and use the same grouping as \citet{wu-dredze-2020-languages}.  In Figure \ref{fig:language_wiki} we present results for two tasks, NER and XNLI.\footnote{Plots for other tasks can be found in the Appendix.} First, we observe that the difference between the model variants is more apparent for NER than for XNLI. This is because NER being a sequence labeling task, the fine-tuned model for NER relies heavily on token-level representations as compared to XNLI which uses the full sequence representation ([CLS] token) for final prediction. This probably leads the sequence labeling models to be more sensitive to the different variants.


Specifically for NER and POS, we observe that languages with small wikipedia sizes are benefited more which suggests that the additional cross-lingual supervision in the form of bilingual dictionary is particularly helpful for low to mid-resource languages. Further, we find that for a given task, the proposed methods perform consistently  across all language families i.e. in most cases the different model variants have a similar rank order with respect to their performance across the language families. This is promising as researchers do not need to pre-train multiple model variants for a given task saving much time and computation effort.


\paragraph{Is the language conditioning layer in \textsc{dict-mlm*}  necessary?}
In Section \S \ref{sec:algorithm}, we discussed two architectural changes to the mBERT model where one of them was adding a language conditioning dense layer for the (dict) MLM prediction task.
Specifically, we concatenated the associated language's embedding with the token's embedding from the final transformer layer  to aid in the cross-lingual synonym prediction.
To evaluate the effect of the conditioning layer, we conduct an ablation study by removing the conditioning layer and report the results on  five downstream tasks for the \textsc{dict-mlm-50} model variant in Table \ref{tab:ablation}.
We find that for NER, removing the conditioning layer causes a significant drop in the  performance as compared to XNLI and MLDOC-H.  For POS and PAWS-X we observe a slight increase in performance upon removing the language conditioning layer. We do note that \textsc{dict-mlm-50}  is the best performing model variant for NER which could suggest that the conditioning layer is indeed providing additional performance gain. However, we leave this for future work to further investigate this observation.

%% file: relatedword.tex
\section{Related Work}
There has been a growing trend towards using explicit alignment objectives to encourage cross-linguality in word representations.  \citet{upadhyay-etal-2016-cross}  show that training word representations with different levels of cross-lingual alignment supervision improves downstream performance on both semantic and syntactic tasks. They use supervision in the form of parallel text with word alignments, an expensive cross-lingual resource, to predict words cross-lingually. They also find that for syntactic tasks such as dependency parsing, a weaker supervision signal in the form of bilingual dictionaries is competitive with other alignment objectives which require parallel text.

Recently, such alignment objectives have also been incorporated with pre-trained models. The most recent work is perhaps by \citet{wu20emnlp} which proposes various methods to align the source and target word representations using parallel text and observe slight improvements over mBERT. However, for a larger capacity model such as XLM-R they do not observe significant gains under an extensive evaluation setup. XLM  \cite{lample2019cross} also uses parallel data with the MLM objective to implicitly encourage cross-lingual alignment. In contrast, our work demonstrates significant improvements over existing works while using only bilingual dictionaries which are more easily available than parallel text.

Bilingual dictionaries have been used for training task-specific models such as in \citet{liu2019attentioninformed} where they use an attention matrix to select the words for dictionary replacement and create code-mixed sentences. Parallel to our work \citet{qin2020cosdaml} also propose to use bilingual dictionaries to construct  code-switched sentences for training downstream models.  \citet{qin2020cosdaml} follow a similar approach as ours of random sampling both word and their synonyms. Our work differs from \citet{qin2020cosdaml, liu2019attentioninformed} in two aspects: 1) we pre-train our model from scratch  whereas \citet{qin2020cosdaml} fine-tune the mBERT model on the multilingual code-switched sentences, and 2) we propose two architectural changes to the pre-trained model to provide additional language signals wherein we add language embeddings to the input layer and a language conditioning dense layer after the transformer stack. 

%% file: conclusion.tex
\section{Conclusions and Future Work}

In this work, we presented  pre-training methods targeted towards improving the cross-lingual representation learning of BERT-based models. We propose using bilingual dictionaries for this purpose since they are a relatively cheap cross-lingual resource and easily available for several languages. We find that our proposed methods outperform existing work  which use stronger cross-lingual supervision in the form of parallel text, on six tasks in the difficult \emph{zero-shot} setting. Furthermore, we find that languages which do not have dictionary resources available are also benefited by our proposed pre-training methods making this method applicable to even endangered languages which might not have dictionaries easily available. 

One limitation of our work is that during the training data generation, the cross-lingual synonyms are selected randomly and out of context which often results in ungrammatical sentences being generated. We plan to address this problem in our future work.
We also plan to expand our methodology to leverage alternative resources such as multilingual word embeddings to further improve the learned representations.  

%% file: appendix.tex
\clearpage
\newpage
\appendix

\section{Appendix}
\label{sec:appendix}
\subsection{Analysis}

\begin{figure}[h]
\centering
\subfigure[POS]{
\label{pos}%
\includegraphics[width=\columnwidth]{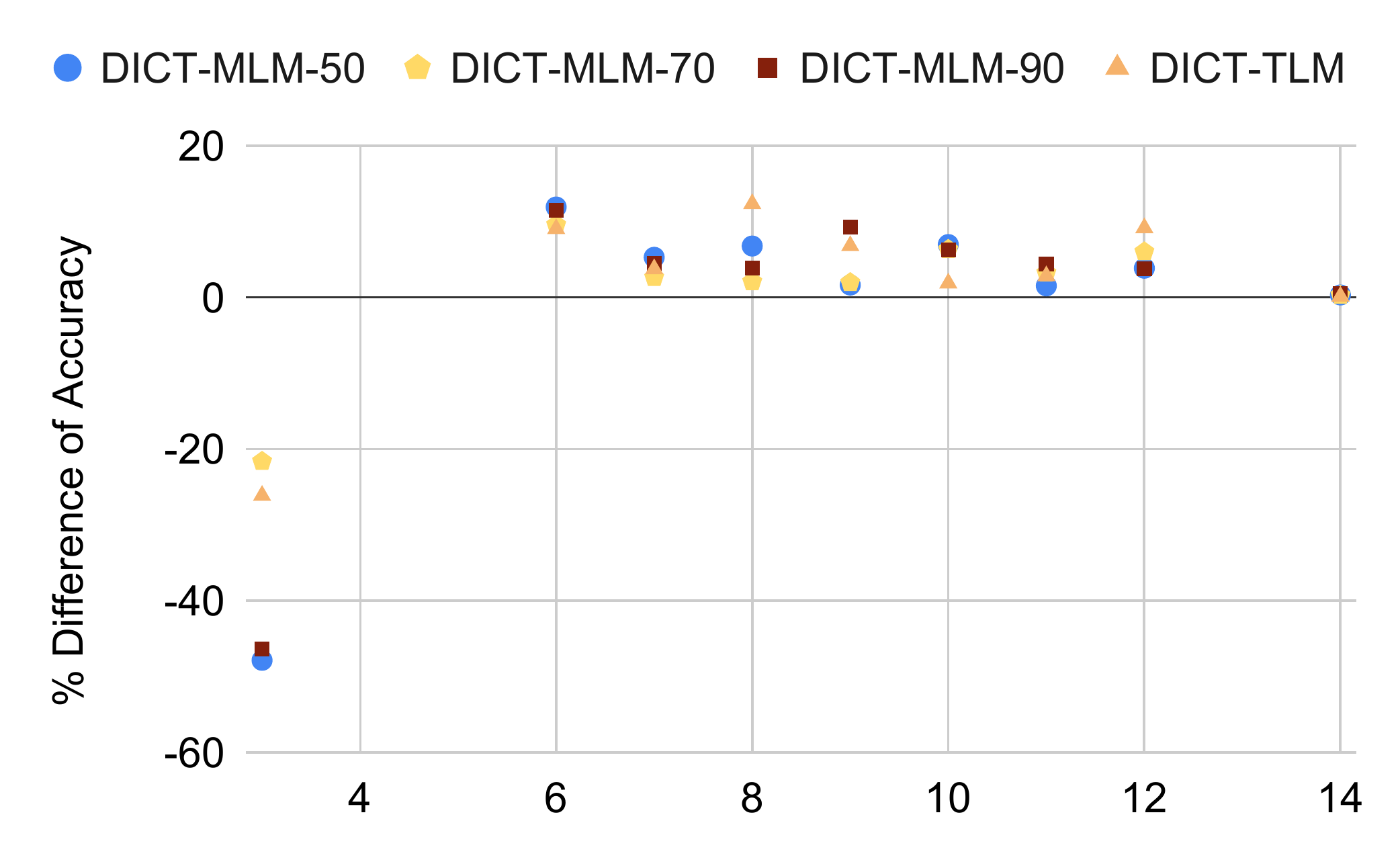}}
~
\subfigure[PAWX-X]{
\label{pawsx}%
\includegraphics[width=\columnwidth]{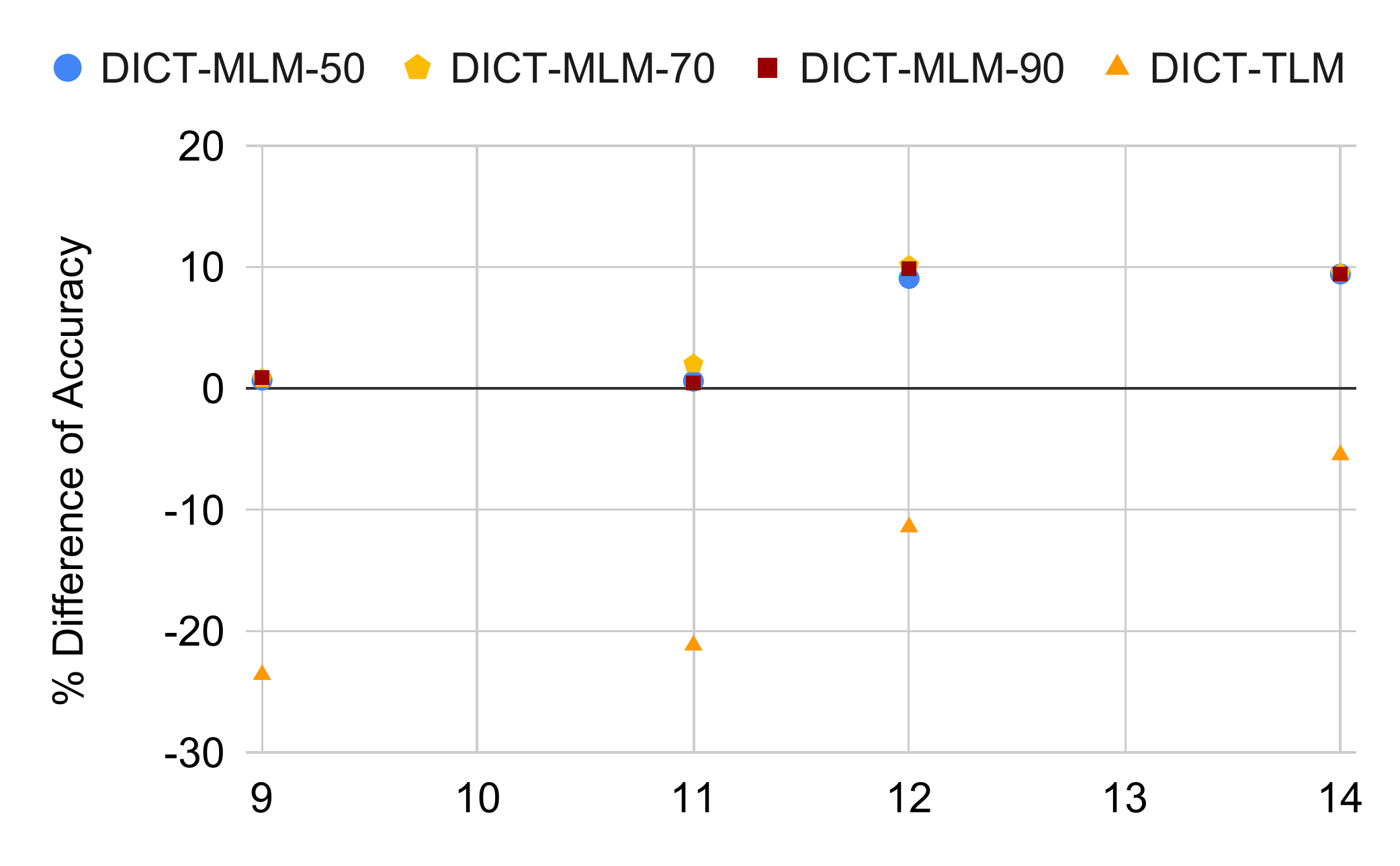}
}
~
\subfigure[MLDOC-H]{
\label{mldoc}%
\includegraphics[width=\columnwidth]{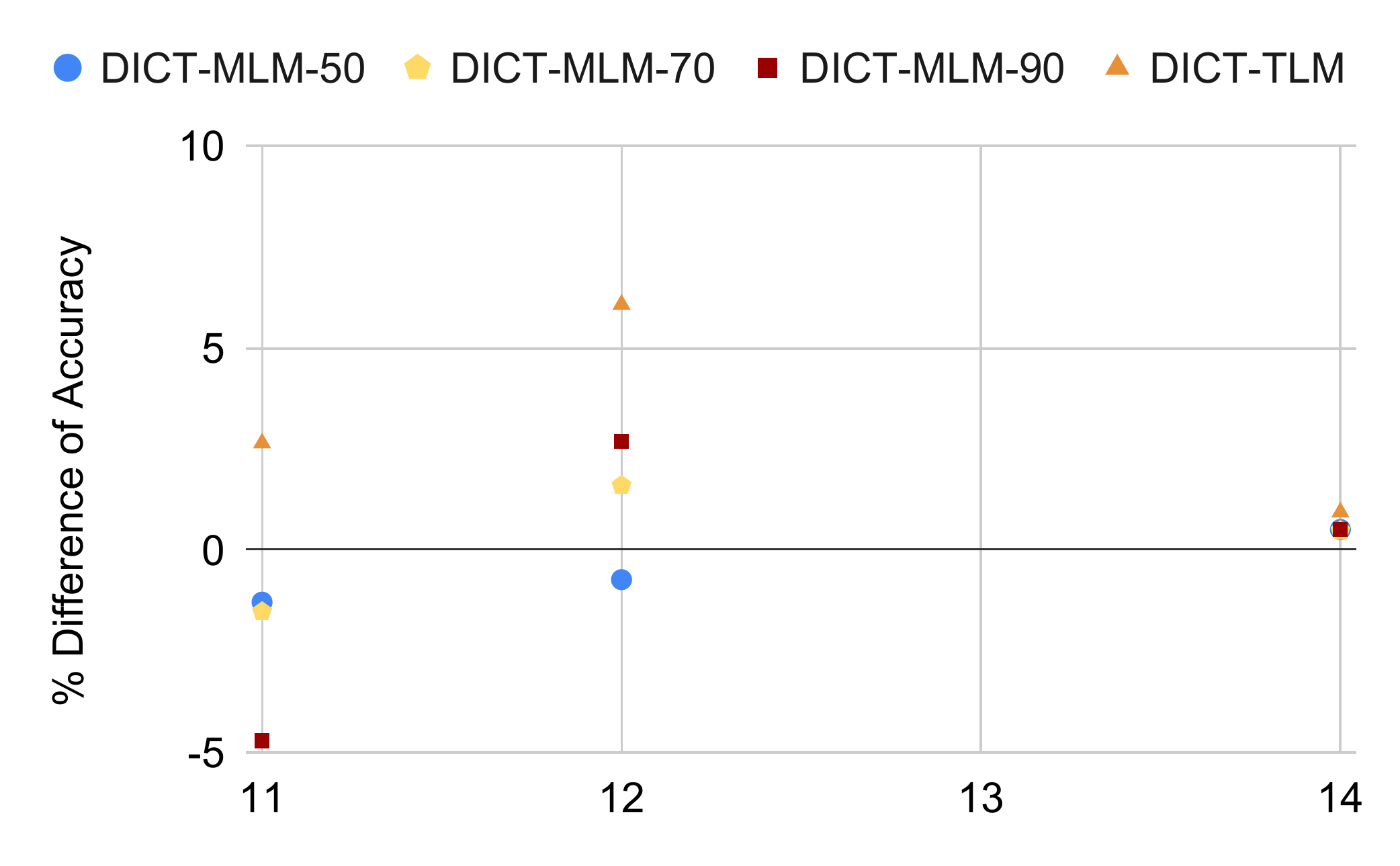}}
~
  \caption{Comparing the (avg.) performance across the different pre-trained models grouped by language family. }%
\label{fig:wiki}%
\end{figure}

\begin{figure*}[h]
\centering
\subfigure[NER]{
\label{pos}%
\includegraphics[width=\columnwidth]{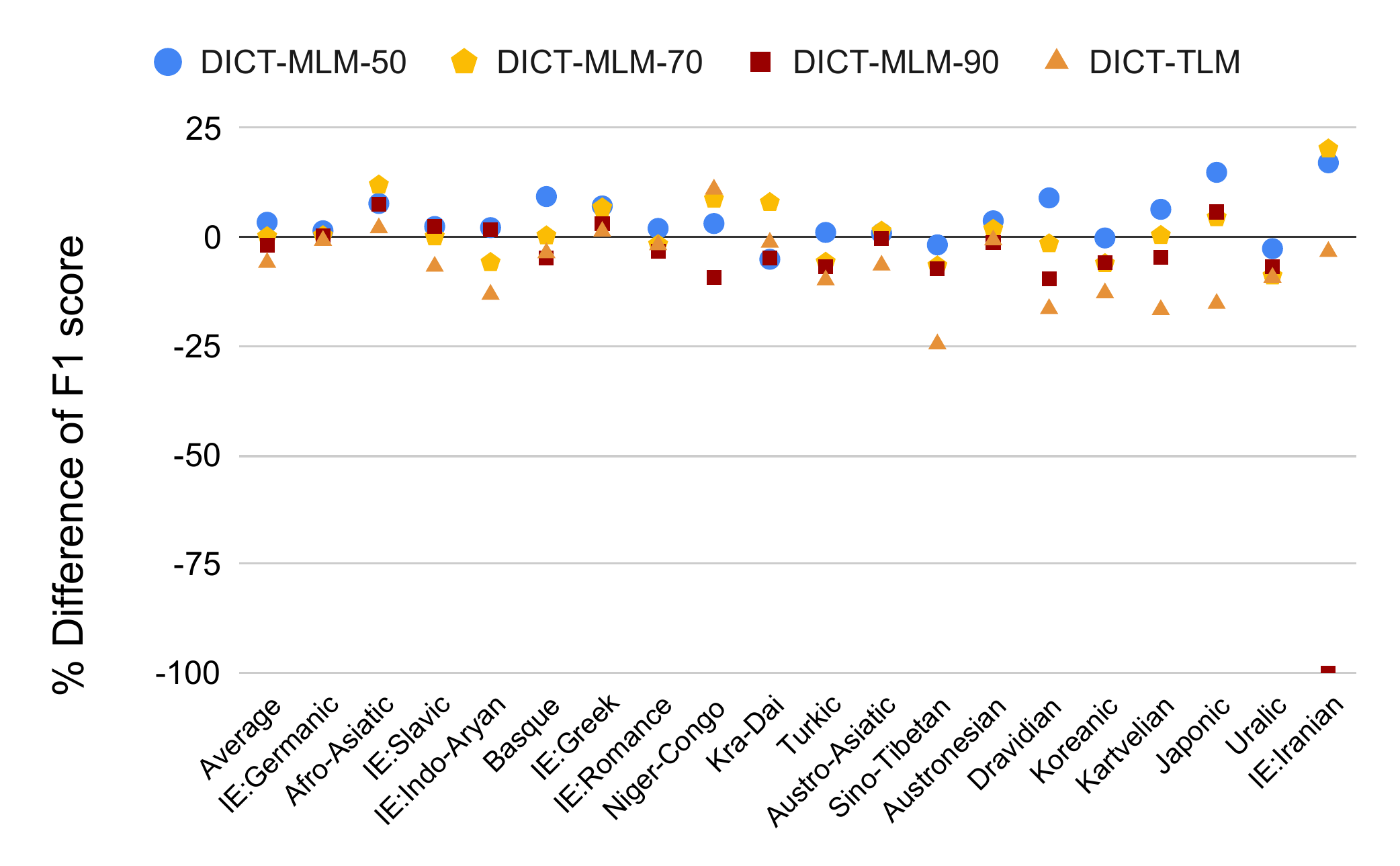}}
~
\subfigure[XNLI]{
\label{pos}%
\includegraphics[width=\columnwidth]{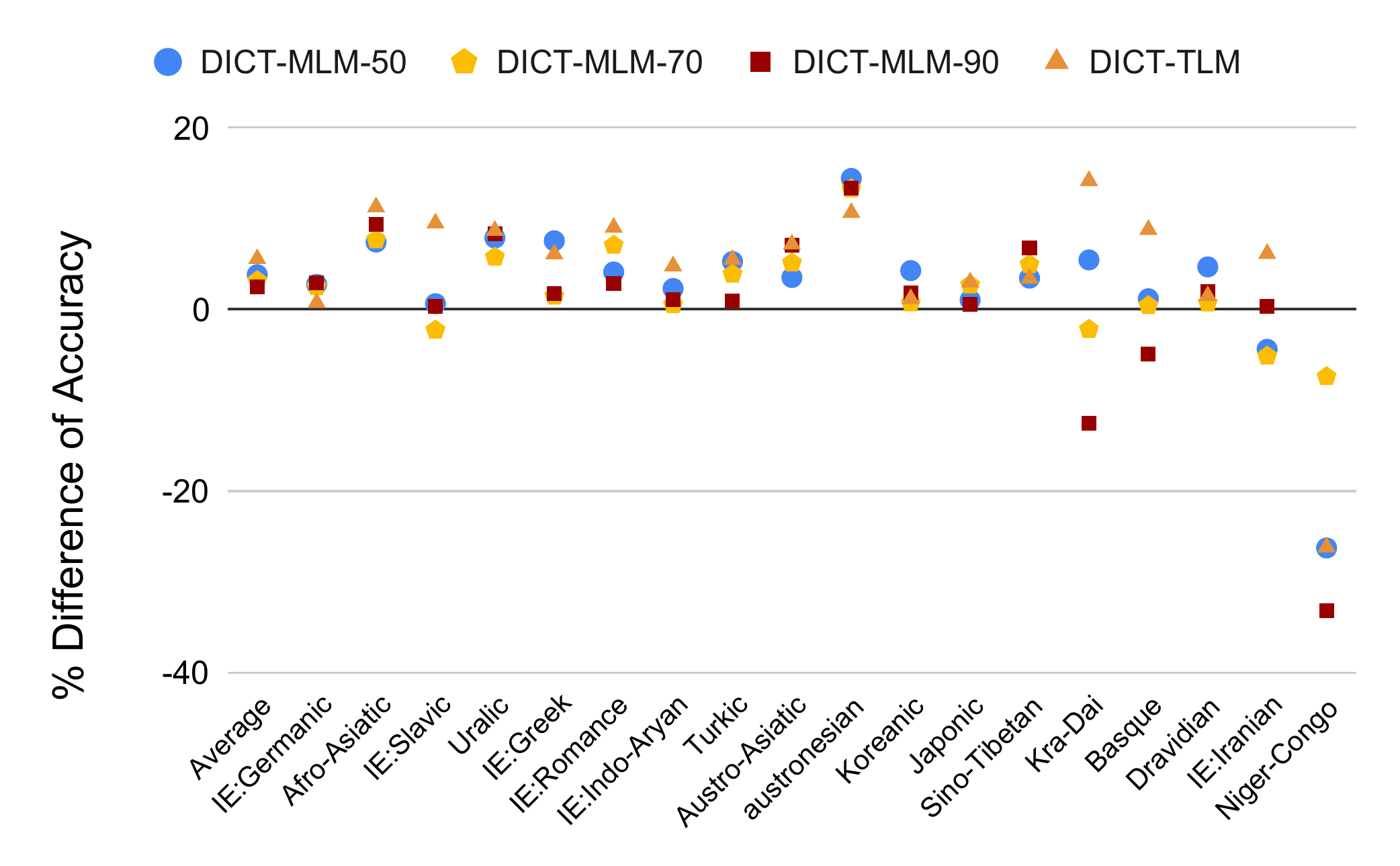}}
~
\subfigure[POS]{
\label{pos}%
\includegraphics[width=\columnwidth]{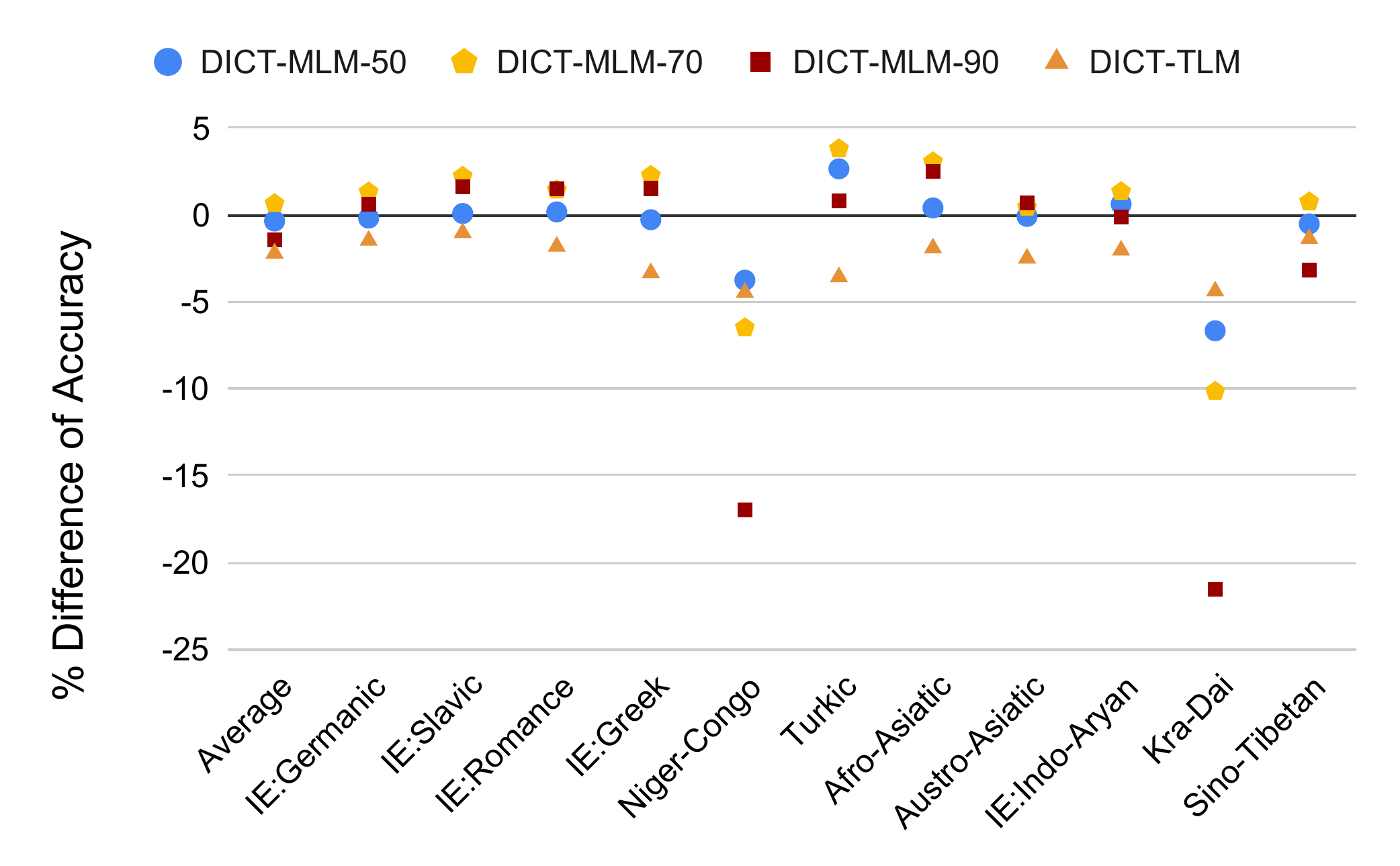}}
~
\subfigure[PAWX-X]{
\label{pawsx}%
\includegraphics[width=\columnwidth]{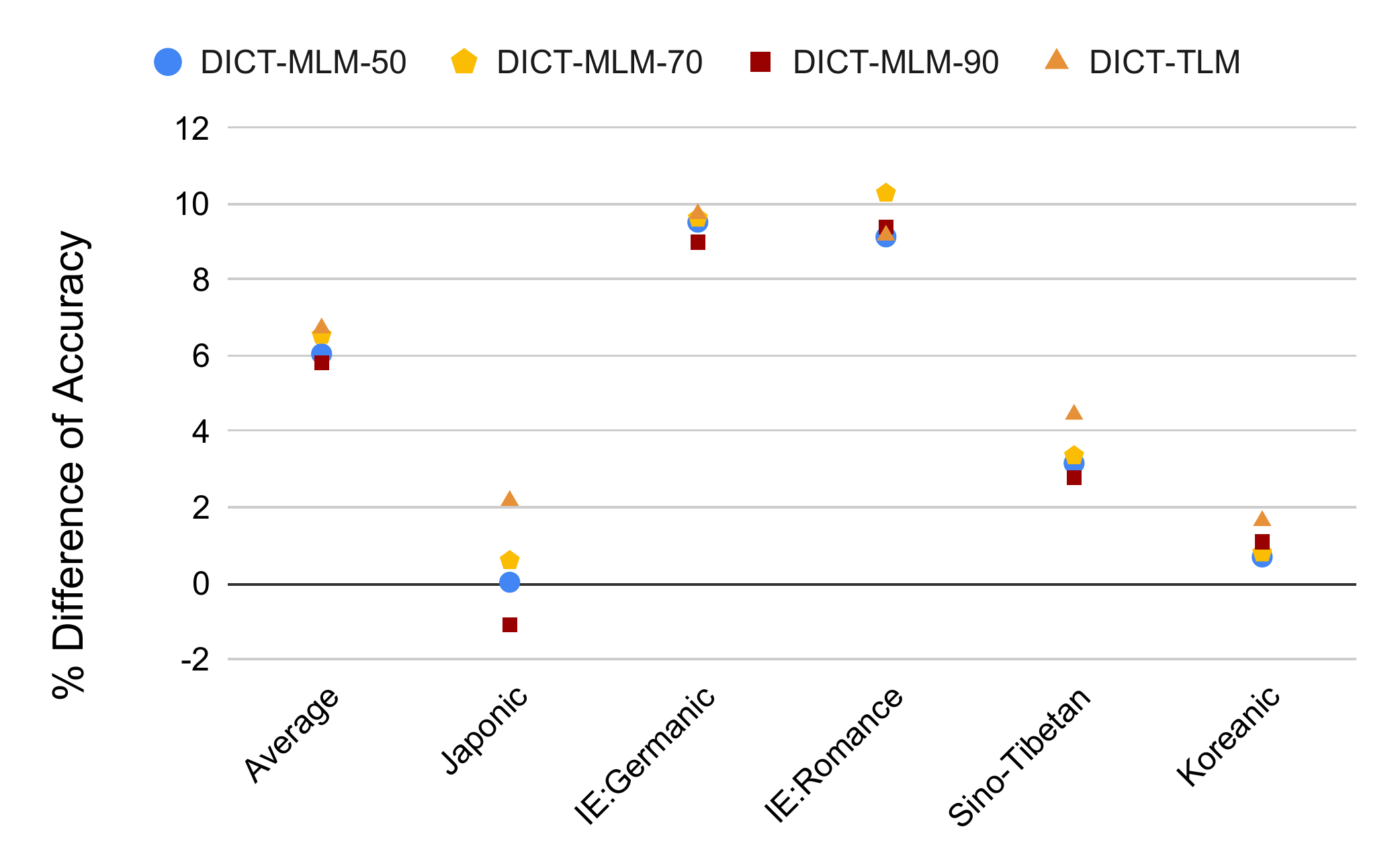}
}
~
\subfigure[MLDOC-H]{
\label{mldoc}%
\includegraphics[width=\columnwidth]{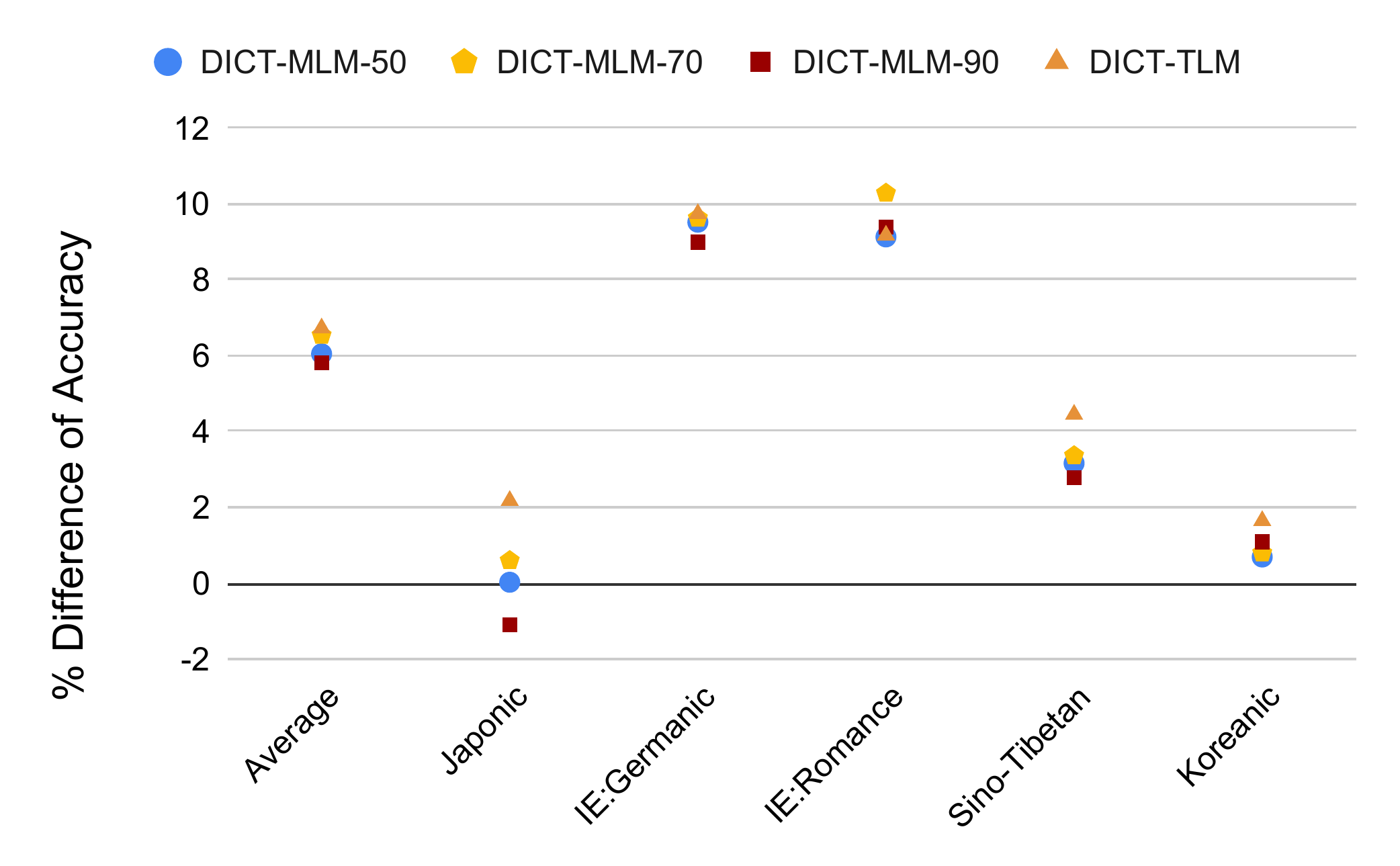}}
~
  \caption{Comparing the (avg.) performance across the different pre-trained models grouped by language family. }%
\label{fig:lang}%
\end{figure*}

We  compare the performance of the model variants across two settings: \emph{language-family} (Figure \ref{fig:lang}) and \emph{wikipedia-size} (Figure \ref{fig:wiki}) and present results for five tasks, grouped by the respective language families and wikipedia sizes. We follow the same language family grouping as XTREME \cite{hu2020xtreme} and for Wikipedia size based grouping we refer to Table 1 in \citet{wu-dredze-2020-languages}.  



